\documentclass[letterpaper, 10 pt, conference]{ieeeconf}  

\IEEEoverridecommandlockouts                              

\usepackage{tikz}
\usepackage{url}
\def\*#1{\mathbf{#1}}

\usepackage{amsmath}
\usepackage{amsfonts}
\usepackage{dsfont}
\usepackage{booktabs} 
\usepackage{multirow}
\usepackage{graphicx}
\usepackage{cleveref}

\usepackage{amssymb}
\usepackage{pifont}
\newcommand{\cmark}{\ding{51}}%

\usepackage{siunitx}

\definecolor{unlabeled}{HTML}{000000}
\definecolor{snow_fall}{HTML}{d52941}
\definecolor{car}{HTML}{6496F5}
\definecolor{road}{HTML}{FF00FF}
\definecolor{other_ground}{HTML}{003f88}
\definecolor{building}{HTML}{FFC800}
\definecolor{vegetation}{HTML}{00AF00}
\definecolor{other_osbtacle}{HTML}{32FFFF}

\title{\LARGE \bf
Energy-based Detection of Adverse Weather Effects in LiDAR Data
}

\author{Aldi Piroli$^{1}$, Vinzenz Dallabetta$^{2}$, Johannes Kopp$^{1}$, Marc Walessa$^{2}$, \\ Daniel Meissner$^{2}$, Klaus Dietmayer$^{1}$
\thanks{$^{1}$ Institute of Measurement, Control, and Microtechnology, Ulm University, Germany {\tt\small \{firstname.lastname\}@uni-ulm.de}}
\thanks{$^{2}$ BMW~AG, Petuelring 130, 80809~Munich,~Germany {\tt\small \{vinzenz.dallabetta, marc.walessa\}@bmw.de} and {\tt\small daniel.da.meissner@bmwgroup.com}}%
}

\newcommand\copyrighttext{%
	\footnotesize \copyright\,2023 IEEE. Personal use of this material is permitted. Permission from IEEE must be obtained for all other uses, in any current or future media, including reprinting/republishing this material for advertising or promotional purposes, creating new collective works, for resale or redistribution to servers or lists, or reuse of any copyrighted component of this work in other works.}%
\newcommand\copyrightnotice{%
	\begin{tikzpicture}[remember picture,overlay]%
	\node[anchor=south,yshift=10pt] at (current page.south) {\fbox{\parbox{\dimexpr\textwidth-2cm}{\copyrighttext}}};%
	\end{tikzpicture}%
	\vspace{-10pt}%
}

\begin{document}

\maketitle
\copyrightnotice
\thispagestyle{empty}
\pagestyle{empty}

\begin{abstract}
Autonomous vehicles rely on LiDAR sensors to perceive the environment.
Adverse weather conditions like rain, snow, and fog negatively affect these sensors, reducing their reliability by introducing unwanted noise in the measurements.
In this work, we tackle this problem by proposing a novel approach for detecting adverse weather effects in LiDAR data.
We reformulate this problem as an outlier detection task and use an energy-based framework to detect outliers in point clouds.
More specifically, our method learns to associate low energy scores with inlier points and high energy scores with outliers allowing for robust detection of adverse weather effects.
In extensive experiments, we show that our method performs better in adverse weather detection and has higher robustness to unseen weather effects than previous state-of-the-art methods.
Furthermore, we show how our method can be used to perform simultaneous outlier detection and semantic segmentation.
Finally, to help expand the research field of LiDAR perception in adverse weather, we release the SemanticSpray dataset, which contains labeled vehicle spray data in highway-like scenarios.
The dataset is available at \url{https://semantic-spray-dataset.github.io}
\end{abstract}

\section{Introduction}
\label{sec:intro}
LiDAR sensors are commonly used in autonomous driving applications alongside cameras and radars.
LiDARs offer precise and rich depth information independently of the lighting conditions, helping an autonomous vehicle to understand and navigate its surroundings. 
The biggest drawback of LiDAR sensors is their sensitivity to adverse weather conditions such as rain, snow, and fog. 
Most perception algorithms developed for LiDAR sensors are trained and tested in favorable weather conditions.
However, their performance is seen to degrade in adverse weather~\cite{walz2021benchmark, mirza2021robustness}, severely reducing their reliability.
With the increasing number of autonomous vehicles on the roads, it is fundamental that they can navigate the environment and safely interact with other road users regardless of weather conditions.

\begin{figure}[t!]
    \centering
        \includegraphics[width=1\columnwidth]{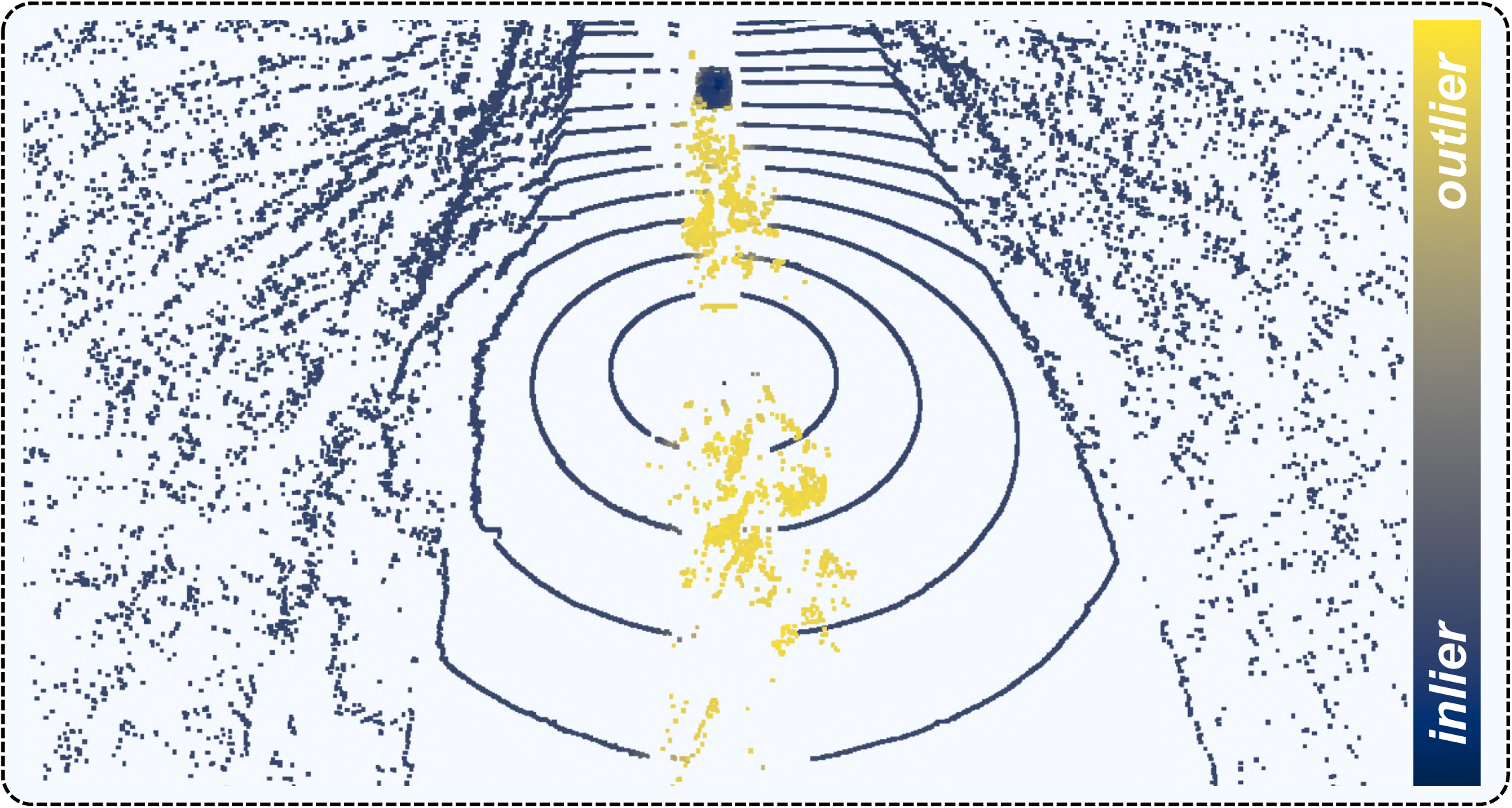}
    \caption{
We propose a method for detecting adverse weather effects in LiDAR data based on energy outlier detection. 
In the figure, we show a scene from the RoadSpray~\cite{road_spray_dataset} dataset, where both a leading vehicle and the ego vehicle are driving on a wet surface, generating trailing spray.
Our model is trained to associate low energy scores (darker colors) with inliers and high energy scores (lighter colors) with outliers, allowing for robust classification of adverse weather effects.
    }
    \label{Fig:teaser}
\end{figure}

In the literature, only a few approaches have been proposed to address this problem~\cite{Piroli2022Robust3O,zhang2021lidar,heinzler2020cnn,sebastian2021rangeweathernet,stanislas2021airborne}. 
Sebastian et al.~\cite{sebastian2021rangeweathernet} classify weather conditions from a LiDAR scan. 
Zhang et al.~\cite{zhang2021lidar} measure the degradation of LiDAR measurements in rainy conditions. 
Both methods only provide information on a scan-wise level, which does not allow the identification of adverse weather effects in the point cloud.
However, detecting which points belong to solid obstacles and which are derived from adverse weather effects is essential for the reliable operation of an autonomous vehicle.  
Heinzler et al.~\cite{heinzler2020cnn} propose a semantic segmentation network to classify adverse weather on a point-wise level. 
Due to the challenge of labeling adverse weather data, they test their model on data collected in a weather chamber, where a static scenario is created and artificial fog and rain are overlayed.
However, real-world scenarios are much more challenging since complex dynamic scenarios can occur under different weather conditions. 

In this work, we address the limitations mentioned above by proposing a method for the point-wise detection of adverse weather conditions.
Instead of directly classifying adverse weather points, we reframe the task as an outlier detection problem.
For this purpose, we adapt the energy-based outlier detection framework proposed in~\cite{liu2020energy} to the 3D point cloud domain.
More specifically, we rewrite the energy score formulation proposed in~\cite{liu2020energy} for the point cloud domain and extend the proposed energy loss function to account for large class imbalances.  
We use this approach to differentiate between inlier points (buildings, vehicles, pedestrians, etc.) and outlier points (spray, rain, snow, fog, etc.). 
Fig.~\ref{Fig:teaser} shows a qualitative result of our approach in detecting spray.
In extensive experiments, we show that our method performs better than previous state-of-the-art methods on real and simulated data.
When training on datasets containing a single adverse weather effect, our approach shows higher robustness to unseen adverse weather effects, making it more applicable to real-world applications where multiple weather effects can occur simultaneously.
Furthermore, our approach allows for the combined semantic segmentation of inlier points and the detection of outlier points.
When we compare our method with a state-of-the-art network for semantic
segmentation~\cite{zhu2021cylindrical}, we see that our solution performs better in detecting points generated by adverse weather effects.

The lack of publicly available labeled data is the primary reason for the limited research on LiDAR in adverse weather.
With this work, we help increase the research opportunities in this field by releasing the SemanticSpray dataset, which provides semantic labels  for the RoadSpray~\cite{road_spray_dataset} dataset.
The dataset contains scenes of vehicles driving on wet surfaces, which can generate a trailing spray corridor. 
This effect is seen to be highly problematic in high-speed scenarios where a large number of spray points are introduced in the LiDAR measurements~\cite{walz2021benchmark}, impeding the sensors’ field of view and, in extreme cases, causing perception systems to fail.

In summary, our main contributions are:
\begin{itemize}
    \item We adapt the energy-based outlier detection framework~\cite{liu2020energy} to detect adverse weather in LiDAR point clouds by formulating the point energy score and extending the energy loss function to account for large class imbalances.
    \item We show that our method outperforms the previous state-of-the-art approaches in detecting adverse weather effects in LiDAR data and has greater robustness to unseen weather effects. 
   \item We show how our method can be adapted to perform both outlier detection and semantic segmentation of the inlier classes using different network architectures. 
   \item We help expand the critical research field of LiDAR perception in adverse weather conditions by  releasing the SemanticSpray dataset. 
\end{itemize}
\section{Related Work}
\label{sec:related_work}
\subsection{Adverse Weather Detection in LiDAR Data}

Walz et al.~\cite{walz2021benchmark} perform a study on the effect of vehicle spray on LiDAR and camera-based object detectors using a vehicle-mounted setup to simulate spray. 
The results show that state-of-the-art detectors are negatively affected by this weather effect, introducing misdetections and ghost objects in the scene.
Zhang et al.~\cite{zhang2021lidar} propose a method for estimating the degree of degradation for LiDAR scans in rainy conditions.
They use an auto-encoder trained with the DeepSAD framework~\cite{ruff2019deep} to output an anomaly score for an input LiDAR scan. 
Sebastian et al.~\cite{sebastian2021rangeweathernet} propose a CNN-based method for classifying weather and road surface conditions in LiDAR data. 
Similar to~\cite{zhang2021lidar}, the method is not meant for classification on a point-wise level but rather on the entire LiDAR scan.   
Heinzler et al.~\cite{heinzler2020cnn} use a lightweight CNN architecture to detect rainfall and fog on a point-wise level.
They train their network on data recorded inside a weather chamber where a static scenario is created using vehicles and mannequins, and then artificial fog and rain are overlayed. 
Although the proposed method performs well on the test data recorded inside the weather chamber, a generalization to dynamic real-world scenarios is not trivial~\cite{Egelhof22a}. 
Stanislas et al.~\cite{stanislas2021airborne} propose a CNN and voxel-based approach for detecting airborne particles in robotic applications. 
Their CNN-based method uses a state-of-the-art CNN architecture developed for image segmentation.
Their voxel-based approach instead uses a fully 3D convolutional backbone, resulting in high inference times. 
Although the primary goal of~\cite{stanislas2021airborne} is to detect smoke and dust, we test their approaches on adverse weather since airborne particles also include snow, rain, fog, and spray. 

A variety of filtering-based methods has also been developed for the detection of outliers in point clouds. 
This includes voxel grid down-sampling, statistical outlier removal (SOR), and radius outlier removal (ROR). 
However, these general-purpose filters perform poorly on adverse weather condition detection~\cite{charron2018noising}.
For this reason, Charron et al.~\cite{charron2018noising} propose the DROR filter, which aims to detect snow points by dynamically adjusting the radius of neighbor search to compensate for the non-uniform density of LiDAR point clouds. 
Kurup et al.~\cite{kurup2021dsor} propose DSOR, which further improves DROR by considering the mean distance between neighbors. 
Overall, filtering-based approaches have a significant advantage over learning-based methods of not requiring training data. 
However, they have high computational complexity and generalize poorly to unmodeled weather effects, limiting their use in real-world applications.

\subsection{Anomaly Detection Methods}
Many approaches have been developed for anomaly detection, mainly for image data. 
The outlier exposure method introduced in~\cite{hendrycks2018deep} proposes to train a network with an auxiliary dataset of outliers, allowing the network to generalize to unseen anomalies. 
Ruff et al.~\cite{ruff2019deep} use a similar approach, redefining the loss function to map inlier data inside a hypersphere and outliers outside it. 
The anomaly score is then derived by computing the distance of the input from the center of the hypersphere.  
Recently, energy-based methods (EBMs) have emerged as the state-of-the-art for anomaly detection both on an image~\cite{liu2020energy} and pixel~\cite{tian2021pixel} level. 
These methods rely on the energy function~\cite{liu2020energy} to map the classification logits to a single real number. 
The network is trained to create an energy gap between inlier and outlier data which can be used for classification.
In our work, we use a similar approach for the point-wise detection of adverse weather effects in LiDAR data.   
Liu et al.~\cite{liu2019deep} propose the abstention learning paradigm, where a model is trained to abstain from making a classification when it is not certain that the input is an inlier.
For this purpose, an additional class is added to the last classification layer of a network, and its output value is used for uncertainty estimation.
Similar to~\cite{tian2021pixel}, we use the idea of an additional class to extend our proposed network.
However, different from~\cite{tian2021pixel}, we do not optimize the abstention learning loss function and instead use an adapted version of  the energy loss function proposed in~\cite{liu2020energy} to detect outliers.
\begin{figure*}[t!]
    \centering
    \includegraphics[width=0.90\textwidth]{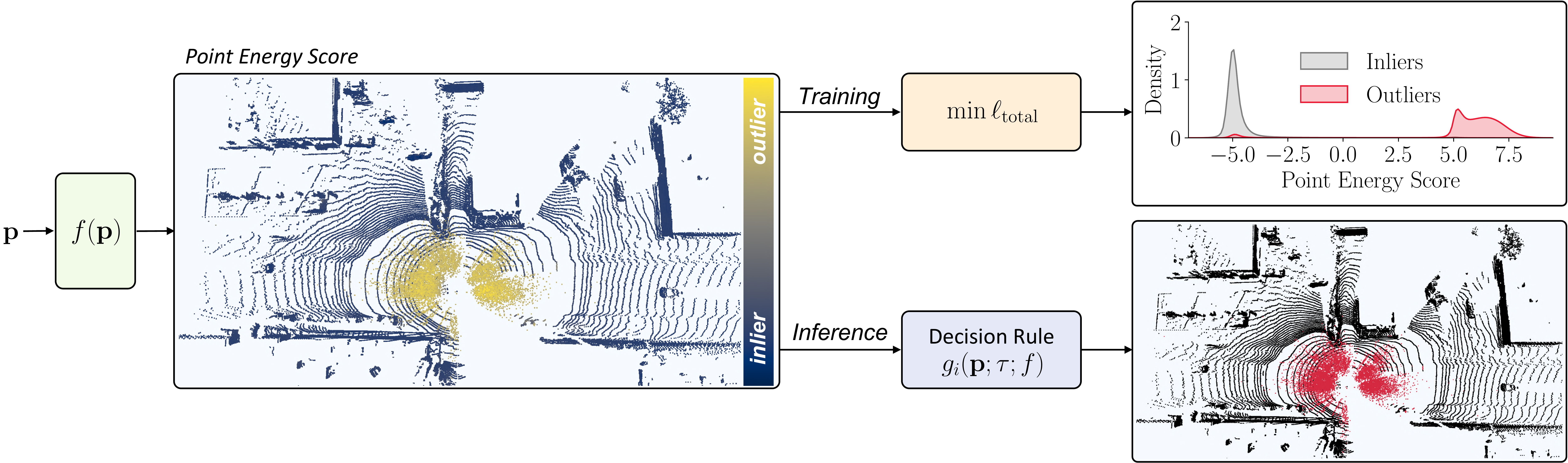}
    \caption{
Given an input point cloud $\*p$, we aim to detect if a point $\*x_i \in \*p$ is caused by adverse weather.
We reframe the problem as an outlier detection task and use the proposed point energy score to detect outlier points.
During training, we minimize the loss function $\ell_\text{total}$~\eqref{eq:total_loss_func}, which results in the model $f(\*p)$ associating a low energy score with inlier points and a high energy score with outliers.
This creates an energy gap between the two categories, which can be used to select a classification threshold $\tau$.
The top-right plot shows an example of the energy gap between inlier and outlier points on the test set of the WADS~\cite{kurup2021dsor} dataset when training $f(\*p)$ on snowy conditions (WADS training set).
During inference, points are classified as outliers (red points in the bottom-right plot) if their energy score is greater than the threshold $\tau$. 
    }
    \label{Fig:method}
\end{figure*}

\section{Method}
\label{sec:method}
This section describes our proposed method for detecting adverse weather points in LiDAR data.
More specifically, our goal is to classify in a binary way if a point in a point cloud is an inlier (not adverse weather) or an outlier (adverse weather). 
In Fig.~\ref{Fig:method}, we show an overview of the approach.

\subsection{Preliminaries}
\textbf{Brief Introduction to EBMs.}
Given a classification model $f(\*x):\mathbb{R}^{D} \rightarrow \mathbb{R}^{K}$, which maps an input $\*x \in \mathbb{R}^D$ to $K$-logits, the \textit{energy function} is defined as:
\begin{align}\label{eq:energy_function}
  E(\*x;f)=- \log \sum_{k \in \{1, \dots, K\}} e^{f(k|\*x)},
\end{align}
with $f(k|\*x)$ being the $k$-th output logit. 
The output of $E(\*x;f):\mathbb{R}^D \rightarrow \mathbb{R}$  represents the \textit{energy score} of the input $\*x$.  
EBMs are characterized by their objective function, which aims to associate low energy scores with inlier inputs and high energy scores with outliers. 
The learned energy gap can then be used to classify inliers and outliers using an appropriate energy threshold \cite{liu2020energy}.

\textbf{Point Energy Score.}
We rewrite the energy function to process 3D inputs on a point-wise level.
Let $\*p = [ \*x_1, \dots, \*x_N ]\in\mathbb{R}^{N \times C}$ be a point cloud with $N$ total points and $\*x_i \in \mathbb{R}^{C}$ a single point with $C$ features.
Given a classification model $f(\*p):\mathbb{R}^{N \times C} \rightarrow \mathbb{R}^{N\times K}$ which outputs $K$-logits for each point $\*x_i \in \*p$, we define the \textit{point energy function} as:
\begin{align}\label{eq:point_wise_energy_function}
  E_i(\*p;f)=- \log \sum_{k \in \{1, \dots, K\}} e^{f_i(k|\*p)}.
\end{align}
The output of the function $E_i(\*p;f):\mathbb{R}^{N \times C} \rightarrow \mathbb{R}$ represents the \textit{point energy score} of $\*x_i$, with $i=\{ 1, \ldots, N\}$.
Here, $f_i(k|\*p)$ is the $k$-th output logit for point $\*x_i$.

\subsection{Point-wise Adverse Weather Detection}
\label{section:point_wise_adverse_weather_detection}
\textbf{Dataset}. 
Given a  point cloud $\*p$, we define the associated point-wise label set as $\*y =\{ y_1, \dots, y_N \}\in\mathbb{N}^{N}$.
We reserve the labels $y_i \in \mathcal{Y}^{in} = \{1, \dots, Y\}$ for inlier classes (buildings, vehicles, pedestrians, etc.) and the label $y_i \in \mathcal{Y}^{out} = \{Y+1\}$ for the outlier class (rain, snow, fog, etc.).

\textbf{Network Architecture}.
Using the abstention learning (AL) framework, we extend the last classification layer of $f(\*p)$ with an additional class.
This class is not used to classify outliers directly but instead allows the model to abstain from classification when it is not confident that a point belongs to one of the $\mathcal{Y}^{in}$ classes. 

\begin{figure}[t!]
    \centering
    \includegraphics[width=0.9\columnwidth]{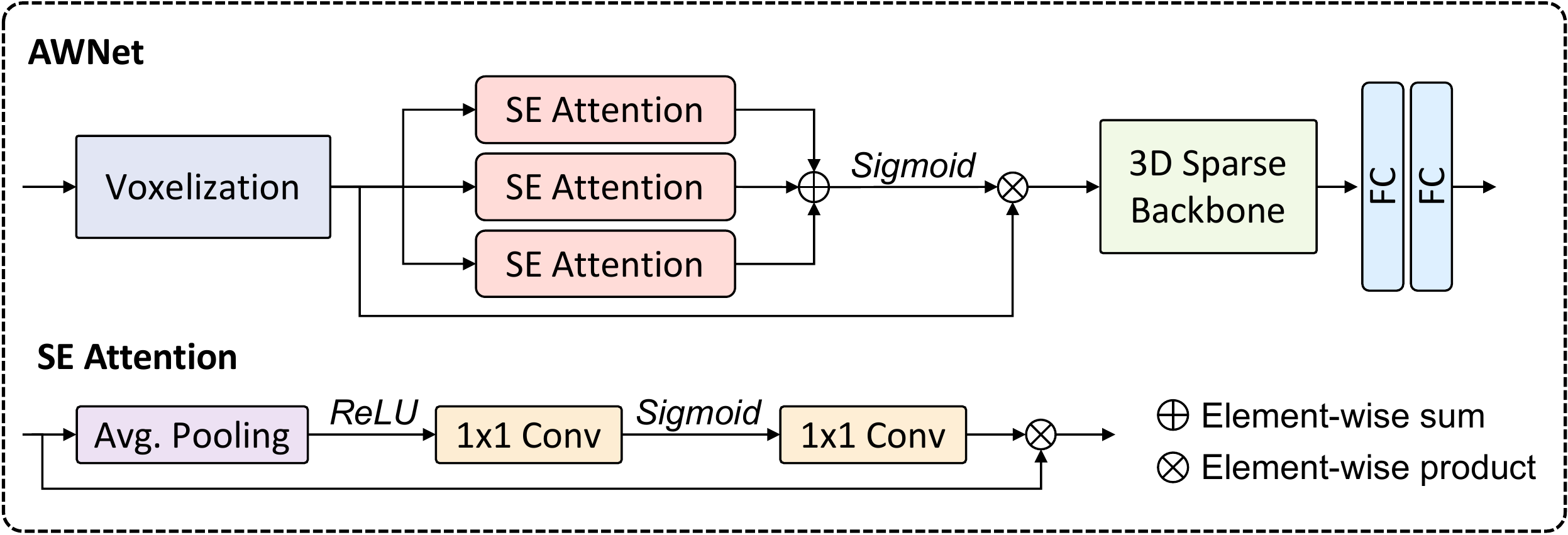}
    \caption{
    AWNet architecture. 
    We use a stack of three squeeze excitation~\cite{hu2018squeeze} (SE) attention layers to weight the channels of each voxel.  
    Then, we extract features using the 3D sparse convolution  proposed in~\cite{shi2020points} and classify each voxel using two fully connected (FC) layers.
    }
    \label{Fig:network_architecture}
\end{figure}

Although our method is not restricted to a particular network architecture, we propose a model for the specific case of adverse weather detection, which we name AdverseWeatherNet (AWNet).
Fig.~\ref{Fig:network_architecture} shows an overview of the architecture.
AWNet first voxelizes the input point cloud and then passes it through a stack of squeeze-excitation attention layers~\cite{hu2018squeeze}.
Each attention layer independently weights the channels of each voxel, and then a combined weight is derived by summing the individual weights.  
This value modulates the voxelized input channels via element-wise multiplication.
The result is then processed using the sparse convolutions backbone presented in~\cite{shi2020points}, which allows for rich feature extraction without the high computational cost of full 3D convolutions.  
Finally, the extracted features are classified using a set of fully connected layers.

\textbf{Loss Function}.
The network $f(\*p)$ is trained using the following loss function:
\begin{align}\label{eq:total_loss_func}
  \ell_\text{total} =  \ell_\text{cls} + \lambda \ell_\text{energy}. 
\end{align}
The first term in~\eqref{eq:total_loss_func} is a semantic segmentation term, which in our case is the standard negative log-likelihood (NLL) loss function defined for an input point cloud $\*p$, and associated point-wise labels $\*y$ as:

\begin{align}\label{eq:loss_func_cls}
 \ell_\text{cls} = \mathbb{E}_{\*x_i \in \*p} \left[ - \log \frac{{f_{i}(y_i|\*p)}}{\sum_{k \in \{1, \dots, K\}} f_i(k|\*p)}  \bigm\vert y_i \in \mathcal{Y}^{in} \right],
\end{align}
with $f_{i}(y_i|\*p)$ being the $y_i$-th output logit.
As shown in~\cite{liu2020energy}, minimizing the NLL results in the model associating a low energy score with the inlier inputs.
Moreover, the loss function allows to learn the semantic segmentation of the inlier classes.
The second term in~\eqref{eq:total_loss_func} is defined for an input point cloud $\*p$ and associated point-wise labels $\*y$ as:
\begin{align}\label{eq:hinge_energy_loss}
\begin{split}
  &\ell_\text{energy}  =  
  \\ &\mathbb{E}_{\*x_i \in \*p}\left[ \omega_{in}^{-1} \big(\max(0, E_i(\*p;f)-m_{in})\big)^2 \bigm\vert y_i \in \mathcal{Y}^{in} \right] +  \\
  &\mathbb{E}_{\*x_i \in \*p}\left[ \omega_{out}^{-1} \big(\max(0, m_{out} - E_i(\*p;f))\big)^2 \bigm\vert y_i \in \mathcal{Y}^{out} \right]
  \end{split}
\end{align}
with:

\begin{align}\label{eq:energy_weighting}
\begin{split}
  \omega_{in} = 1 + \sum_{y_i \in \*y} \mathds{1}_{y_i \in  \mathcal{Y}^{in}}, \;\;  \omega_{out} = 1+ \sum_{y_i \in \*y} \mathds{1}_{y_i \in  \mathcal{Y}^{out}}.
\end{split}
\end{align}
The $\ell_\text{energy}$~\eqref{eq:hinge_energy_loss} term is a hinge loss function that penalizes inlier energies greater than the margin parameter $m_{in} \in \mathbb{R}$ and outlier energies lower than $m_{out}\in \mathbb{R}$~\cite{liu2020energy}. 
This creates an energy gap that can be used to define an adequate threshold to differentiate between inlier and outlier points. 
The task of point-wise anomaly detection in the case of adverse weather conditions can be an unbalanced problem.
Even in severe precipitations, the number of non-adverse weather effect points can largely outnumber adverse weather ones. 
This large unbalance causes the $\ell_\text{energy}$~\eqref{eq:hinge_energy_loss} term to saturate, resulting in insufficient supervision during training for the outlier energy association. 
We address this problem using the terms described in~\eqref{eq:energy_weighting}, which weight the energy loss terms proportionally to the number of inlier and outlier points present in the point cloud. 
The parameter $\lambda\in \mathbb{R}$ is used to weight the loss term $\ell_\text{energy}$~\eqref{eq:hinge_energy_loss}.

\subsection{Training and Inference}
\textbf{Training}.
Like other energy-based methods, we use a fine-tuning approach for training~\cite{liu2020energy, tian2021pixel}.
Given a model trained on a dataset containing only inlier points, we first extend its final classification layer using the AL framework described in Section~\ref{section:point_wise_adverse_weather_detection}.
Then, we fine-tune the model weights on a dataset that contains both inlier and outlier points using the loss function  $\ell_\text{total}$~\eqref{eq:total_loss_func}.
This approach has the advantage of being efficient since it does not require the entire retraining of the network.

\textbf{Inference}.
At inference time, our model returns the energy score for each point in the point cloud. 
An appropriate energy threshold  $\tau\in \mathbb{R}$ can be chosen depending on the application at hand. 
For example, in autonomous driving applications, $\tau$ can be chosen so that a high fraction (e.g., $95\%$) of inlier points is correctly classified. 
The threshold $\tau$ is common for all points during inference.
The decision rule $g_i$ for a point $\*x_i \in \*p$ can then be formulated as:
\begin{align}
\label{eq:decision_rule}
g_i(\*p; \tau; f) =
\begin{cases}
\text{\textit{inlier}} & \quad \text{if } E_i(\*p;f) \leq \tau, \\
\text{\textit{outlier}}  & \quad \text{else}.
\end{cases}
\end{align}

\section{The SemanticSpray Dataset}
\label{sec:semantic_road_spray_dataset}
Within this work, we also release the SemanticSpray dataset.
The data is based on the recently released RoadSpray~\cite{road_spray_dataset} dataset, which performs large-scale testing of the effect of vehicle spray on LiDAR, camera, and radar sensors.
The RoadSpray dataset provides unlabeled scenes of vehicles driving on wet a surface at different speeds in highway-like scenarios.
For the sensor setup and additional information on the experiments, we refer the reader to the original publication [3].
To create the SemanticSpray dataset, we manually label each point in a LiDAR scan as one of three categories: \textit{background}, \textit{vehicle}, and \textit{spray}. 
All static objects (road, vegetation, buildings, etc.) are labeled as \textit{background}.
The moving vehicles in the scene are labeled as \textit{vehicle}.
The \textit{spray} class contains the spray generated by the ego vehicle and other moving vehicles in the scene. 
In total, we provide semantic labels for $16565$ dynamic scenes, with approximately $6.23\cdot10^6$ \textit{background}, $4.84\cdot10^4$ \textit{vehicle} and $2.15\cdot10^4$ \textit{spray} points.
The dataset is available for download at \url{https://semantic-spray-dataset.github.io}. The dataset toolkit is instead available at \url{https://github.com/aldipiroli/semantic_spray_dataset}.
\begin{table}[t!]
    \centering
    
    \caption{Evaluation results of adverse weather detection when training and testing on the same weather effect. 
    Values are in percentage. 
    $\uparrow$ means higher values are better, and $\downarrow$ that lower values are better.}
    \resizebox{0.9\columnwidth}{!}{%
\begin{tabular}{@{}llccc@{}}
            \toprule
            \textbf{Method}                                                 & $\mathcal{D}_\text{test}$ & \textbf{AUROC}~$\uparrow$ & \textbf{AUPR}~$\uparrow$ & \textbf{FPR95}~$\downarrow$ \\ \midrule
\multirow{5}{*}{Particle-UNet~\cite{stanislas2021airborne}}                & WADS             & 97.18          & 97.17          & 12.62                \\
                                     & SemanticSpray      & 99.69          & 99.24          & 0.31                 \\
                                     & NuScenes-Fog    & 99.98          & 99.98          & 0.02                 \\
                                     & DENSE     & 99.61          & 98.96          & 0.68                 \\ \cmidrule(l){2-5} 
                                     & \textbf{average} & 99.11          & \textbf{98.84} & 3.41                 \\ \midrule
\multirow{5}{*}{Particle-VoxelNet~\cite{stanislas2021airborne}}          & WADS             & 98.71          & 98.48          & 2.02                 \\
                                     & SemanticSpray      & 99.16          & 98.96          & 0.70                 \\
                                     & NuScenes-Fog    & 99.88          & 99.87          & 0.15                 \\
                                     & DENSE     & 92.02          & 82.01          & 37.57                \\ \cmidrule(l){2-5} 
                                     & \textbf{average} & 97.44          & 94.83          & 10.11                \\ \midrule
\multirow{5}{*}{WeatherNet~\cite{heinzler2020cnn}}        & WADS             & 97.40          & 96.68          & 10.59                \\
                                     & SemanticSpray      & 99.46          & 95.47          & 2.88                 \\
                                     & NuScenes-Fog    & 99.89          & 99.60          & 0.04                 \\
                                     & DENSE     & 99.49          & 98.90          & 1.94                 \\ \cmidrule(l){2-5} 
                                     & \textbf{average} & 99.06          & 97.66          & 3.86                 \\ \midrule
\multirow{5}{*}{AWNet (ours)} & WADS             & 98.26          & 96.89          & 1.24                 \\
                                     & SemanticSpray      & 99.84          & 99.25          & 0.02                 \\
                                     & NuScenes-Fog    & 99.99          & 99.98          & 0.01                 \\
                                     & DENSE     & 99.27          & 97.55          & 0.77                 \\ \cmidrule(l){2-5} 
                                     & \textbf{average} & \textbf{99.34} & 98.42          & \textbf{0.51}        \\ \bottomrule
\end{tabular}
}
\label{tab:dtest_same_as_dtrain}
\end{table}

\section{Experiments}
\label{sec:experiments}
\subsection{Experiment Setup}
\textbf{Baselines.} We evaluate our approach against the other methods for adverse weather~\cite{heinzler2020cnn, charron2018noising,kurup2021dsor} and airborne particle detection in LiDAR data~\cite{stanislas2021airborne}.
Each learning-based baseline is trained to classify inlier (non-adverse weather), and outlier (adverse weather) points using the default parameters. 
Previous methods evaluate only on a single dataset which usually includes a single adverse weather condition. 
However, this testing approach is unrealistic since, in many real-world scenarios, multiple weather effects can
occur simultaneously, i.e., snowfall and wet surface spray.
Therefore, we test each method on the adverse weather effects used during training and on unseen ones.

\textbf{Datasets.} 
We evaluate the performance in a variety of different adverse weather conditions. 
SemanticSpray is the dataset presented in Section~\ref{sec:semantic_road_spray_dataset}, which contains spray effects in highway-like scenarios.
We use $7898$ scans for training and $8667$ for testing.
The WADS~\cite{kurup2021dsor} dataset was recorded in snowy conditions while driving in urban environments. 
We use $1011$ scans for training and $918$ for testing\footnote{Used training/test splits: $\{13$, $14$, $17$, $20$, 
$23$,
$26$,
$30$,
$34$,
$35$,
$36\}$ / $\{$$11$,
$12$,
$15$,
$16$,
$18$,
$22$,
$24$,
$28$,
$37$,
$76$$\}$.
}. 
The DENSE~\cite{heinzler2020cnn} dataset contains static scenarios inside a weather chamber where artificial rainfall and fog are generated.  
We use the official train ($61900$ scans), and test ($19787$ scans) splits. 
To test our proposed method on more complex scenarios under foggy conditions, we use the simulation method proposed in~\cite{HahnerICCV21} to augment with fog the NuScenes dataset~\cite{caesar2020nuscenes} (NuScenes-Fog).
The dataset has $27449$ scans for training and $6019$ for testing.
Since our main goal is to differentiate between inliers 
and outliers, unless otherwise stated, we consider all the possible non-adverse weather classes (e.g., vehicle, pedestrian, building, etc.) as the inlier class. 
Similarly, we consider all of the adverse weather classes (e.g., snow, fog, and spray) as the outlier class. 
\begin{table*}[t!]
    \centering
    \caption{Evaluation results of adverse weather detection when training on an adverse weather effect and testing on an unseen (during training) weather effect. 
    For each result, $\mathcal{D}_\text{train}$ represents the training set, and  $\mathcal{D}_\text{test}$ the test set. Values are in percentage. $\uparrow$ means higher values are better, and $\downarrow$ that lower values are better.}
    \resizebox{0.53\textwidth}{!}{%
\begin{tabular}{@{}lllccc@{}}
            \toprule
            \textbf{Method}                                                 & $\mathcal{D}_\text{train}$ & $\mathcal{D}_\text{test}$ & \textbf{AUROC}~$\uparrow$ & \textbf{AUPR}~$\uparrow$ & \textbf{FPR95}~$\downarrow$ \\ \midrule
\multirow{5}{*}{Particle-UNet~\cite{stanislas2021airborne}}                & WADS               & SemanticSpray     & 84.33 & 52.14 & 59.59 \\
                                     & SemanticSpray        & WADS            & 74.57 & 63.75 & 70.17 \\
                                     & NuScenes-Fog      & SemanticSpray     & 61.94 & 55.40 & 83.22 \\
                                     & NuScenes-Fog      & WADS            & 93.30 & 92.74 & 35.58 \\ \cmidrule(l){2-6} 
                                     & \multicolumn{2}{l}{\textbf{average}} & 78.53 & 66.01 & 62.14 \\ \midrule
\multirow{5}{*}{Particle-VoxelNet~\cite{stanislas2021airborne}}          & WADS               & SemanticSpray     & 96.13 & 69.49 & 18.15 \\
                                     & SemanticSpray        & WADS            & 93.02 & 90.14 & 23.10 \\
                                     & NuScenes-Fog      & SemanticSpray     & 87.54 & 82.09 & 28.57 \\
                                     & NuScenes-Fog      & WADS            & 95.32 & 95.06 & 10.07 \\ \cmidrule(l){2-6} 
                                     & \multicolumn{2}{l}{\textbf{average}} & 93.00 & 84.19 & 19.97 \\ \midrule
\multirow{5}{*}{WeatherNet~\cite{heinzler2020cnn}}        & WADS               & SemanticSpray     & 94.40 & 71.72 & 33.94 \\
                                     & SemanticSpray        & WADS            & 81.58 & 68.53 & 58.64 \\
                                     & NuScenes-Fog      & SemanticSpray     & 85.17 & 57.34 & 65.12 \\
                                     & NuScenes-Fog      & WADS            & 88.83 & 87.00 & 58.99 \\ \cmidrule(l){2-6} 
                                     & \multicolumn{2}{l}{\textbf{average}} & 87.49 & 71.15 & 54.17 \\ \midrule
\multirow{5}{*}{AWNet (ours)} & WADS               & SemanticSpray     & 99.30 & 93.61 & 0.86  \\
                                     & SemanticSpray        & WADS            & 96.74 & 91.93 & 7.19  \\
                                     & NuScenes-Fog      & SemanticSpray     & 96.29 & 70.52 & 21.24 \\
                                     & NuScenes-Fog      & WADS            & 97.35 & 95.26 & 4.52  \\ \cmidrule(l){2-6}
               & \multicolumn{2}{l}{\textbf{average}} & \textbf{97.42}                     & \textbf{87.83}                  & \textbf{8.45}                            \\ \bottomrule
\end{tabular}
}
\label{tab:dtest_different_than_dtrain}
\end{table*}

\textbf{Evaluation Metrics.} 
Following prior work~\cite{hendrycks2016baseline, hendrycks2018deep,  liu2020energy, tian2021pixel}, we evaluate the outlier detection task using the area under receiver operating characteristics (AUROC), area under the precision-recall curve (AUPR) and the false positive rate at $95\%$ true positive rate (FPR95). 
To compare against statistical filters, we use precision and recall metrics.
Additionally, we use the intersection over union (IoU) and the mean IoU (mIoU) to evaluate semantic segmentation performances.

\textbf{Implementation Details.}
We pretrain AWNet using the NuScenes~\cite{caesar2020nuscenes} dataset on the foreground (vehicles, pedestrians, cyclists, etc.) and background class (everything else).
We train for $30$ epochs on $20\%$ of the training set using the NLL loss, Adam optimizer~\cite{Kingma2014AdamAM}, and constant learning rate of $10^{-4}$.
For the energy-based fine-tuning, we train for a maximum of $20$ epochs using Adam optimizer and training parameters $m_\text{in} = -5$, $m_\text{out} = 5$ and $\lambda=0.1$.
When training on the SemanticSpray and DENSE datasets, we set $\omega_{in}$ and $\omega_{out}$ equal to $1$ since there is a lower imbalance between inlier and outlier points.
We implement AWNet with voxel size of $[ 0.1, 0.1, 0.2]$ $\si{\metre}$ in the $x$, $y$ and $z$ direction respectively.
A full description of the sparse backbone parameters is given in~\cite{shi2020points}.
The final classification layer comprises two fully connected layers of size $256$.
\begin{table}[t!]
    \centering
    \caption{Evaluation of our proposed method against statistical filters. Our method is trained and tested on the same dataset. Values are in percentage.  $\uparrow$ means higher values are better, and $\downarrow$ that lower values are better.}
    \resizebox{0.75\columnwidth}{!}{%
\begin{tabular}{@{}llcc@{}}
            \toprule
            \textbf{Method}                                &
            \multicolumn{1}{l}{$\mathcal{D}_\text{test}$}  &
            \multicolumn{1}{l}{\textbf{Precision}~$\uparrow$}  &
            \multicolumn{1}{l}{\textbf{Recall}~$\uparrow$}                                                                               \\ \midrule
\multirow{5}{*}{DROR~\cite{charron2018noising}}        & WADS             & 85.48              & 88.11           \\
                                     & SemanticSpray      & 50.68              & 64.24           \\
                                     & NuScenes-Fog    & 70.00              & 84.09           \\
                                     & DENSE     & 57.92              & 67.65           \\ \cmidrule(l){2-4} 
                                     & \textbf{average} & 66.02              & 76.02           \\ \midrule
\multirow{5}{*}{DSOR~\cite{kurup2021dsor}}        & WADS             & 76.92              & 88.85           \\
                                     & SemanticSpray      & 51.57              & 56.58           \\
                                     & NuScenes-Fog    & 75.47              & 89.72           \\
                                     & DENSE     & 64.56              & 69.76           \\ \cmidrule(l){2-4} 
                                     & \textbf{average} & 67.13              & 76.23           \\ \midrule
\multirow{5}{*}{AWNet (ours)} & WADS             & 91.11              & 95.45           \\
                                     & SemanticSpray      & 87.49              & 99.32           \\
                                     & NuScenes-Fog    & 99.22              & 99.84           \\
                                     & DENSE     & 91.43              & 95.73           \\ \cmidrule(l){2-4} 
                                     & \textbf{average} & \textbf{92.31}     & \textbf{97.58}  \\ \bottomrule
\end{tabular}
}
\label{tab:filter_comparison}
\end{table}

\begin{table*}[t!]
    \centering
    \caption{Semantic segmentation evaluation when training and testing on WADS. Values are in percentage. }
    \resizebox{0.75\textwidth}{!}{%
\begin{tabular}{@{}lccccccccc@{}}
\toprule
\multirow{2}{*}{\textbf{Model}} & \multicolumn{8}{c}{\textbf{IoU}}                                                & \multirow{2}{*}{\textbf{mIoU}} \\
 & \textit{unlabeled} & \textit{car} & \textit{road} & \textit{other-ground} & \textit{building} & \textit{vegetation} & \textit{other-obstacle} & \textit{snowfall} &  \\ \midrule
Cylinder3D~\cite{zhu2021cylindrical} &
  29.76 &
  63.33 &
  51.31 &
  \textbf{62.34} &
  \textbf{68.41} &
  53.32 &
  15.41 &
  78.75 &
  52.83 \\
 Cylinder3D-E~\cite{zhu2021cylindrical} + ours &
\textbf{32.58} & 
\textbf{65.29} & 
47.76 & 
61.71 & 
66.84 & 
52.09 & 
17.12 & 
79.48 & 
\textbf{52.86} \\
AWNet (ours) &
12.44 &
56.43 &
\textbf{58.06} &
60.77 &
67.64 &
\textbf{60.06} &
\textbf{17.32} &
\textbf{87.03} &
52.47 \\ \bottomrule
\end{tabular}
}
\label{tab:semantic_segmentation}
\end{table*}

\subsection{Results}
\begin{figure}[t!]
    \centering
\includegraphics[width=0.90\columnwidth]{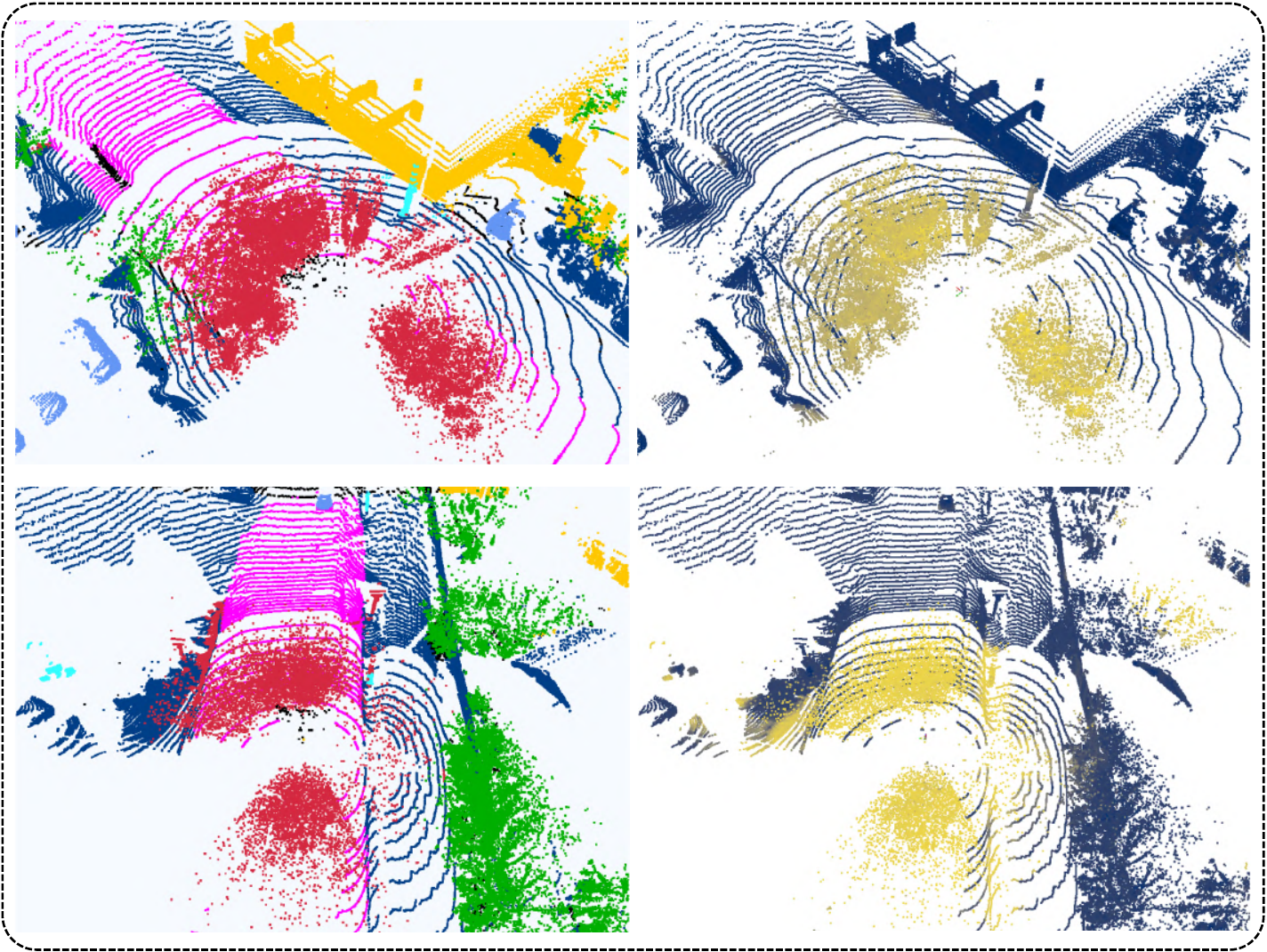}
    \caption{
    The top row shows an example of AWNet trained and tested on the WADS dataset.
    The bottom row shows instead an example of AWNet trained on SemanticSpray and tested on WADS.
    On the left column, we show the semantic ground truth labels, whereas on the right, the output point energy score where lighter colors represent inlier scores and brighter colors outliers. 
    The ground truth classes are represented with the following colors: $\color{unlabeled}{\bullet}$~\textit{unlabeled},  
    $\color{car}{\bullet}$~\textit{car},  
    $\color{road}{\bullet}$~\textit{road},  
    $\color{other_ground}{\bullet}$~\textit{other-ground},  
    $\color{building}{\bullet}$~\textit{building},  
    $\color{vegetation}{\bullet}$~\textit{vegetation},  
    $\color{other_osbtacle}{\bullet}$~\textit{other-obstacle},  
    $\color{snow_fall}{\bullet}$~\textit{snowfall}.
    }
    \label{Fig:edge_cases}
\end{figure}

\textbf{Training and Testing on the Same Weather Effect.} 
We start the evaluation by comparing the performance of AWNet against the other methods by training and testing in the same weather conditions. 
We report the results in Tab.~\ref{tab:dtest_same_as_dtrain}.
Our method achieves the best average performance across all tested datasets in terms of AUROC and FPR95, improving the latter by $2.9\%$ points compared to the second-best method (Particle-UNet).
This last result shows that our method is well suited for safety-critical applications like autonomous driving, where a high percentage of adverse weather effects needs to be detected without having a large number of false positives.
Fig.~\ref{Fig:teaser} and Fig.~\ref{Fig:edge_cases}-top show qualitative results of the point energy score output of AWNet trained and tested on vehicle spray and snow respectively.

\begin{figure}[t!]
    \centering
\includegraphics[width=0.85\columnwidth]{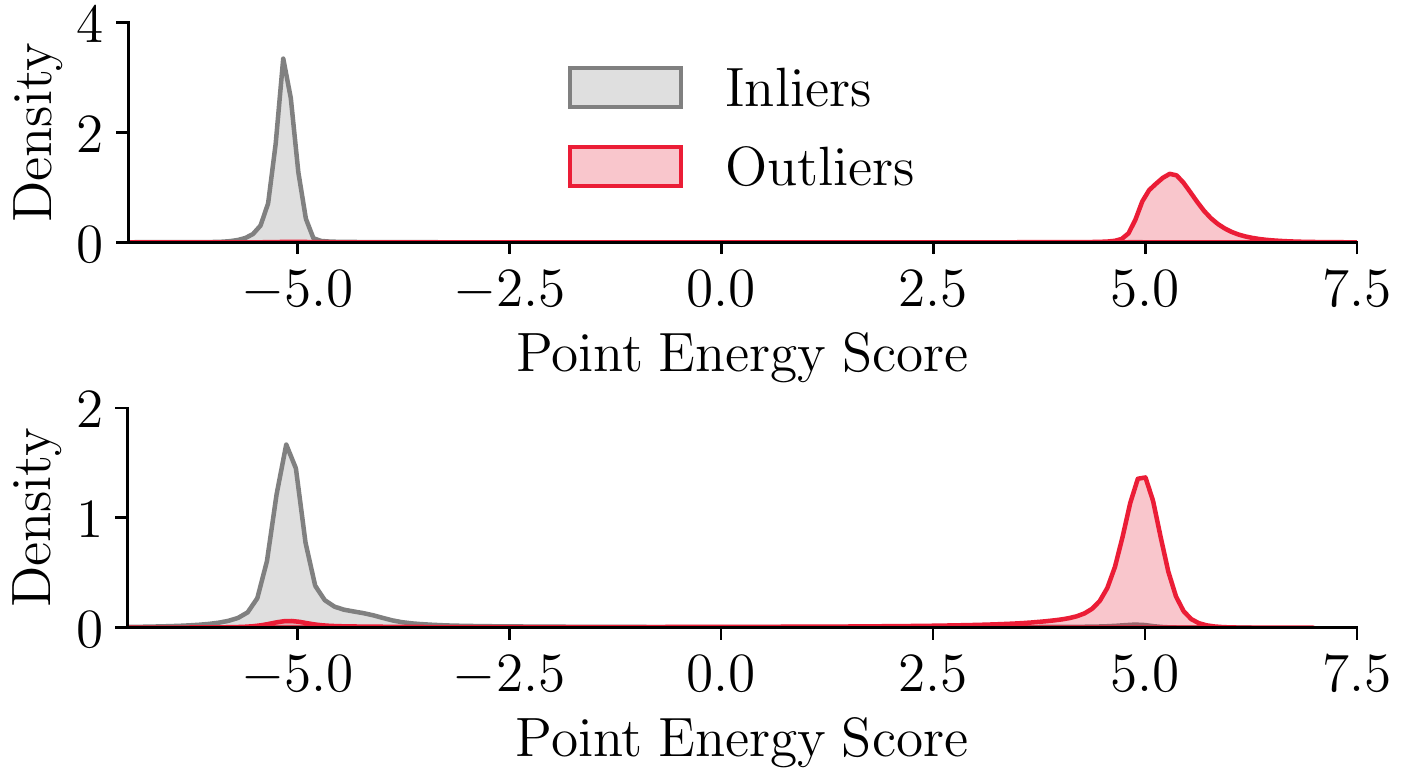}
    \caption{
    Point energy score distributions of AWNet trained on the SemanticSpray dataset and tested on the SemanticSpray (top) and WADS (bottom) datasets. 
    Although the network is trained only on spray data, the model associates similar energy scores to snow and spray points.
    This highlights the robustness of our method to unseen weather effects since a common classification threshold $\tau$ can be chosen to classify both  effects.
    }
    \label{Fig:score_distribution}
\end{figure}

\textbf{Training and Testing on Different Weather Conditions.}
In Tab.~\ref{tab:dtest_different_than_dtrain}, we report the results of the methods trained on a single weather condition $\mathcal{D}_\text{train}$ and tested on unseen (during training) weather conditions $\mathcal{D}_\text{test}$.
We see that AWNet outperforms the other methods by a large margin across all metrics.
When training on snowy conditions (WADS) and testing on spray (SemanticSpray), AWNet achieves an improvement of $3.17\%$~AUROC, $24.12\%$~AUPR and $17.29\%$~FPR95 points against the second best network (Particle-VoxelNet).
We see similar performances when training on spray and testing on snowfall.
Although snowfall and spray are very different effects,
our method can detect both as adverse weather, even when
training with only one of the weather effects. 
This result shows that compared to the other approaches, the proposed point-wise energy-based outlier detection has a higher robustness to unseen adverse weather effects.
This can also be observed in Fig.~\ref{Fig:score_distribution}, where AWNet associates similar energy scores for both seen and unseen weather effects.
This property is desirable for real-world applications where is not possible to cover all possible weather variations during training.
A qualitative example of AWNet trained on SemanticSpray and tested on WADS is reported in Fig.~\ref{Fig:edge_cases}-bottom.
We also see that both voxel-based approaches (AWNet and Particle-VoxelNet) perform better than the CNN-based methods when training and testing on different sensors.
For example, the WADS dataset was recorded using a $64$ layers LiDAR whereas the SemanticSpray dataset with a $32$ layers sensor.
CNN-based methods rely on a range image projection which is dependent on both the sensor resolution and sampling rate.
In contrast, voxel-based methods are less affected by the sensor properties since they use a fixed voxel  partitioning.
In our experiments, we also tested the generalizability of models trained on the DENSE dataset. However, similar to what~\cite{Egelhof22a} reported, the restricted field of view and the internal structure of the fog chamber yields poor performances on outdoor datasets with $360^{\circ}$ field of view.

\textbf{Training With Simulated Data.}
All four methods perform well when training and testing with simulated fog data (NuScenes-Fog).
When testing on snowfall (WADS), AWNet improves by $5.55\%$ FPR95 points against the second-best method (Particle-VoxelNet).
However, when testing on the SemanticSpray dataset, AWNet has lower performances than Particle-VoxelNet in terms of AUPR, suggesting that there is still a margin for improvement.

\textbf{Comparison Against Statistical Filters.} 
In Table~\ref{tab:filter_comparison}, we show the comparison between our proposed method and the statistical filters.
Our method largely outperforms both filters in all datasets.
DROR and DSOR perform well on snow (design task) and fog detection.
When testing on spray, we see lower performances instead.
Unlike snowfall and fog, spray is detected in dense clusters, making the use of statistical properties like the number of neighbors less effective for point-wise filtering.  

\textbf{Simultaneous Semantic Segmentation and Outlier Detection.} 
In many applications, in addition to outlier detection, the segmentation of the inlier class might also be needed. 
Our method is well suited for this task since, in addition to differentiating between inliers and outliers, we also learn the semantic segmentation of the inlier classes using $\ell_\text{cls}$~\eqref{eq:loss_func_cls}.
We test the performances of AWNet against Cylinder3D~\cite{zhu2021cylindrical}, a state-of-the-art semantic segmentation network, on the WADS dataset.
Both networks are trained from scratch using only the WADS dataset.
We train AWNet for $30$ epochs using only $\ell_\text{cls}$~\eqref{eq:loss_func_cls} and then fine tune for $20$ epochs using $\ell_\text{total}$~\eqref{eq:total_loss_func} with parameters $m_{in}=-4.5$, $m_{out}=0$ and  $\lambda=0.1$. 
During inference, we determine the semantic class of a point by choosing the maximum softmax logit among the $K+1$ outputs. 
In addition, we classify as \textit{snowfall} all points $\*x_i$ which have $E(\*x_i;f) > \tau$, with $\tau = -0.25$.
We train Cylinder3D using the default parameters described in~\cite{zhu2021cylindrical}.
Additionally, we train the same network using our proposed approach (Cylinder3D-E).
First, we train for $30$~epochs with the default parameters, then we fine-tune for $20$ epochs using both the original loss function and $\ell_\text{energy}$~\eqref{eq:total_loss_func} with parameters $m_{in}=-5$, $m_{out}=5$ and $\lambda=1$.
For inference, we use the same approach described for AWNet with $\tau = 4.5$.  
During training and evaluation, we also use the \textit{unlabeled} class since we observe that a small portion of the \textit{snowfall} points is incorrectly labeled as such.
We show the results in Tab.~\ref{tab:semantic_segmentation}.
Overall, Cylinder3D achieves a higher mIoU across the inlier classes than AWNet.
However, in the adverse weather class (\textit{snowfall}), AWNet outperforms Cylinder3D by $8.28\%$ IoU points.
Using our fine-tuning approach (Cylinder3D-E) improves the performances of Cylinder3D on the outlier class (\textit{snowfall}) by $0.73\%$ IoU points, showing that our method can be used with different network architectures.

\textbf{Inference Time.}
We test the inference time of AWNet and the other methods on the SemanticSpray dataset using an NVIDIA RTX 2080 Ti.
To process a single LiDAR scan AWNet takes on average $15.37$ $\si{\milli\second}$, Particle-VoxelNet $409.14$ $\si{\milli\second}$, Particle-UNet $4.32$ $\si{\milli\second}$, WeatherNet $4.18$ $\si{\milli\second}$, DROR $199.27$ $\si{\milli\second}$ and DSOR $71.50$ $\si{\milli\second}$.
Although both CNN methods have faster inference, AWNet is still real-time capable given the $10\text{-}20$ $\si{\hertz}$ acquisition frequency of common LiDAR sensors.
Furthermore, compared to the previous voxel-based state-of-the-art approach (Particle-VoxelNet), AWNet has more than $26$-times faster inference time.

\textbf{Discussion.}
Our approach demonstrates promising results in distinguishing between inliers and outliers while training and testing under similar weather conditions. 
Additionally, it offers good generalization performance for detecting unseen weather effects. 
However, our method has some limitations.
For instance, when the point distribution of inliers and outliers is similar (e.g., snow and vegetation in Fig.~\ref{Fig:edge_cases}-top), some inlier points incorrectly receive a high point energy score. 
This issue is further accentuated when testing under unseen weather conditions (Fig.~\ref{Fig:edge_cases}-bottom).

\begin{table}[t]
    \centering
    \caption{Ablation studies of our proposed method when training on WADS. Values are in percentage. $\uparrow$ means higher values are better, and $\downarrow$ that lower values are better.}
    \resizebox{1\columnwidth}{!}{%
        \begin{tabular}{@{}lccccccc@{}}
            \toprule
            $\mathcal{D}_\text{test}$                                                &
            \textbf{\begin{tabular}[c]{@{}c@{}}Pretrained\\ Model\end{tabular}}      &
            \textbf{\begin{tabular}[c]{@{}c@{}}Weighted \\ Energy Loss\end{tabular}} &
            \textbf{AL}                                                              &
            \textbf{\begin{tabular}[c]{@{}c@{}}SE \\ Attention\end{tabular}}         &
            \textbf{AUROC $\uparrow$}                                                &
            \textbf{AUPR $\uparrow$}                                                 &
            \textbf{FPR95 $\downarrow$}                                                                                                                                    \\ \midrule
            \multirow{4}{*}{WADS}                                                    & -      & -      & \cmark & \cmark & 97.97          & 96.75          & 1.33          \\
                                                                                     & \cmark & -      & \cmark & \cmark & 98.23          & \textbf{97.05} & \textbf{1.22} \\
                                                                                     & \cmark & \cmark & -      & \cmark & 98.08          & 96.54          & 1.40          \\
                                                                                     & \cmark & \cmark & \cmark & -      & 98.13          & 96.67          & 1.28          \\     
                                                                                     & \cmark & \cmark & \cmark & \cmark & \textbf{98.26} & 96.89          & 1.24          \\ \midrule
            \multirow{4}{*}{SemanticSpray}                                           & -      & -      & \cmark & \cmark & 98.11          & 85.88          & 5.69          \\
                                                                                     & \cmark & -      & \cmark & \cmark & 98.84          & 93.42          & 1.39          \\
                                                                                     & \cmark & \cmark & -      & \cmark & 98.96          & 87.82          & 3.36          \\
                                                                                     & \cmark & \cmark & \cmark & -      & 98.60          & 90.89          & 2.53           \\ 
                                                                                     & \cmark & \cmark & \cmark & \cmark & \textbf{99.30} & \textbf{93.61} & \textbf{0.86} \\ \bottomrule
        \end{tabular}
    }
    \label{tab:ablation_studies}
\end{table}
 
\subsection{Ablation Studies}
\textbf{Components Contribution.}
Tab.~\ref{tab:ablation_studies} shows the contribution of each component used when training AWNet on the WADS dataset.
When we pretrain AWNet on inlier data, we see a small improvement in snowfall detection.
When testing the same model on spray (unseen during training), we see an improvement of $0.73\%$~AUROC, $7.54\%$~AUPR, and $4.30\%$~FPR95 points instead.
This can be associated with the pretrained model already having an internal representation of the inlier class, allowing for a better generalization of the outlier class during training.  
This result is important since most LiDAR perception datasets are recorded in good weather conditions.
Our method shows that this large amount of data can be leveraged to improve the detection of adverse weather without the cost of additional labeled data. 
When we compare the results of AWNet trained from scratch on WADS and tested on SemanticSpray to Particle-VoxelNet, we still see an improvement of $1.98\%$~AUROC, $16.39\%$~AUPR and $12.46\%$~FPR95 points, showing that even without pretraining our method has better robustness to unseen weather effects. 
The point-wise energy loss weighting term described in~\eqref{eq:energy_weighting} further improves the generalizability performances of AWNet when testing on spray.
In our experiments, we observed that in datasets where a large imbalance between inliers and outliers exists (WADS and NuScenes-Fog), the energy weighting term helps the model consistently converge during training. 
Finally, we see that both the AL framework and the SE Attention modules increase performance when testing on the WADS and SemanticSpray datasets.

\textbf{Different Network Architectures.}
In Tab.~\ref{tab:ablation_different_architectures} we test the performance of our proposed  method using different network architectures.
All models are trained from scratch, using the proposed loss function~\eqref{eq:total_loss_func} and class-wise weighting  described in~\eqref{eq:energy_weighting}.
We set $\lambda=0.01$ and margin parameters $m_{in}=-5$, $m_{out}=5$ in all experiments.
The results indicate that our approach can improve performance when evaluating on known and unknown adverse weather conditions. 
For instance, our method improves Particle-UNet's FPR95 score on WADS by $2.8\%$ points. 
Similarly, Particle-VoxelNet's FPR95 score on SemanticSpray improves by $13.12\%$ points. 
The performance gains can be attributed to the energy loss function, which provides additional supervision during training.
By penalizing energy outputs in the range $[m_{in}, m_{out}]$, the network leans a more distinguishable representation between inlier and outliers points.
\begin{table}[t]
    \centering
    \caption{ Application of our proposed loss function to different network architectures. All models are trained from scratch on the WADS dataset. Values are in percentage. $\uparrow$ means higher values are better, and $\downarrow$ that lower values are better.}
    \resizebox{0.9\columnwidth}{!}{%
        \begin{tabular}{@{}llccc@{}}
            \toprule
            \textbf{Method}                                & $\mathcal{D}_\text{test}$      & \multicolumn{1}{l}{\textbf{AUROC} $\uparrow$} & \multicolumn{1}{l}{\textbf{AUPR} $\uparrow$} & \multicolumn{1}{l}{\textbf{FPR95} $\downarrow$} \\ \midrule
            Particle-UNet~\cite{stanislas2021airborne}     & \multirow{3}{*}{WADS}          & 97.62                                         & 97.09                                        & 9.82                                            \\
            Particle-VoxelNet~\cite{stanislas2021airborne} &                                & 98.72                                         & 98.35                                        & 2.19                                            \\
            WeatherNet~\cite{heinzler2020cnn}              &                                & 97.51                                         & 96.43                                        & 8.66                                            \\ \midrule
            Particle-UNet~\cite{stanislas2021airborne}     & \multirow{3}{*}{SemanticSpray} & 87.43                                         & 52.56                                        & 53.37                                           \\
            Particle-VoxelNet~\cite{stanislas2021airborne} &                                & 98.48                                         & 88.48                                        & 5.03                                            \\
            WeatherNet~\cite{heinzler2020cnn}              &                                & 93.77                                         & 72.15                                        & 31.60                                           \\ \bottomrule
        \end{tabular}
    }
    \label{tab:ablation_different_architectures}
\end{table}
\section{Conclusion} 
\label{sec:conclusion}
In this work, we propose a new method for detecting adverse weather effects in LiDAR point clouds.
We reframe the task as an outlier detection problem and use the recently proposed energy-based outlier detection framework to robustly detect adverse weather points.
Extensive experiments on datasets containing spray, rain, snow, and fog show that our method performs better than previous state-of-the-art methods.
Furthermore, our proposed method has higher robustness to unseen weather effects, increasing its applicability to real-world applications. 
Finally, we contribute to the expansion of the critical research field of LiDAR perception in adverse weather conditions by releasing the SemanticSpray dataset, which contains labeled scenes of vehicle  spray data in highway-like scenarios.

\bibliographystyle{IEEEtran}
\bibliography{mybib}

\end{document}